\pdfoutput=1

\documentclass[11pt]{article}

\usepackage[final]{acl}

\usepackage{times}
\usepackage{latexsym}

\usepackage[T1]{fontenc}

\usepackage[utf8]{inputenc}

\usepackage{microtype}

\usepackage{inconsolata}

\usepackage{graphicx}
\usepackage{soul}

\usepackage{amsmath,amssymb} 
\usepackage{algorithm}  
\usepackage{algorithmicx}
\usepackage{algpseudocode} 
\usepackage{graphicx}
\usepackage{multirow}
\usepackage{subfigure}
\usepackage{comment}
\usepackage{booktabs}
\usepackage{stackengine}
\usepackage{xcolor,colortbl}
\usepackage{amsthm}     
\usepackage{caption}
\usepackage{wrapfig}
\usepackage{arydshln}
\usepackage[most]{tcolorbox}
\usepackage{lipsum}
\usepackage{color, colortbl}
\definecolor{Gray}{gray}{0.85}

\makeatletter
\newenvironment{breakablealgorithm}
  {
   \begin{center}
     \refstepcounter{algorithm}
     \hrule height.8pt depth0pt \kern2pt
     \renewcommand{\caption}[2][\relax]{
       {\raggedright\textbf{\ALG@name~\thealgorithm} ##2\par}%
       \ifx\relax##1\relax 
         \addcontentsline{loa}{algorithm}{\protect\numberline{\thealgorithm}##2}%
       \else 
         \addcontentsline{loa}{algorithm}{\protect\numberline{\thealgorithm}##1}%
       \fi
       \kern2pt\hrule\kern2pt
     }
  }{
     \kern2pt\hrule\relax
   \end{center}
  }
\makeatother

\title{${\text{S}^2\text{-MAD}}$: Breaking the Token Barrier to Enhance Multi-Agent Debate Efficiency}

\author{
 \textbf{Yuting Zeng\textsuperscript{1,2}},
 \textbf{Weizhe Huang\textsuperscript{1}},
 \textbf{Lei Jiang\textsuperscript{1}},
 \textbf{Tongxuan Liu\textsuperscript{1,2*}},
 \textbf{Xitai Jin\textsuperscript{3}},
 \\
 \textbf{Chen Tianying Tiana\textsuperscript{4}},
 \textbf{Jing Li\textsuperscript{1*}},
 \textbf{Xiaohua Xu\textsuperscript{1*}}
\\
 \textsuperscript{1}University of Science and Technology of China,
 \textsuperscript{2}JD.com,
 \\
 \textsuperscript{3}Harbin Institute of Technology,
 \textsuperscript{4}National University of Singapore
\\
  \small{
  {$\{$yuting$\_$zeng, hwz871982879, jianglei0510, tongxuan.ltx$\}$@mail.ustc.edu.cn},
 }
 \\
 \small{
 {tianachen@u.nus.edu}, {$\{$lj, xiaohuaxu$\}$@ustc.edu.cn}, {2023212227@stu.hit.edu.cn}
 }
}

\begin{document}
\maketitle
\begin{abstract}
Large language models (LLMs) have demonstrated remarkable capabilities across various natural language processing (NLP) scenarios, but they still face challenges when handling complex arithmetic and logical reasoning tasks. While Chain-Of-Thought (CoT) reasoning, self-consistency (SC) and self-correction strategies have attempted to guide models in sequential, multi-step reasoning, Multi-agent Debate (MAD) has emerged as a viable approach for enhancing the reasoning capabilities of LLMs. By increasing both the number of agents and the frequency of debates, the performance of LLMs improves significantly. However, this strategy results in a significant increase in token costs, presenting a barrier to scalability. To address this challenge, we introduce a novel sparsification strategy designed to reduce token costs within MAD. This approach minimizes ineffective exchanges of information and unproductive discussions among agents, thereby enhancing the overall efficiency of the debate process. We conduct comparative experiments on multiple datasets across various models, demonstrating that our approach significantly reduces the token costs in MAD to a considerable extent. Specifically, compared to MAD, our approach achieves an impressive reduction of up to 94.5\% in token costs while maintaining performance degradation below 2.0\%.
\end{abstract}

\renewcommand{\thefootnote}{\fnsymbol{footnote}}
\footnotetext[1]{Corresponding authors.}

\section{Introduction}
\label{sec:introduction}
Large language models (LLMs) have shown exceptional capabilities across a variety of natural language processing (NLP) tasks \citep{openai2024gpt4,brown2020language,bubeck2023sparks,radford2018improving,radford2019language,touvron2023llama,touvron2023llama2,anil2023palm,chowdhery2023palm}. 
However, even the most advanced LLMs exhibit limitations in complex mathematical reasoning and logical inference scenarios \citep{liu2023evaluating}. 

 \begin{figure*}[t]
    \centering
    \includegraphics[width=1\linewidth]{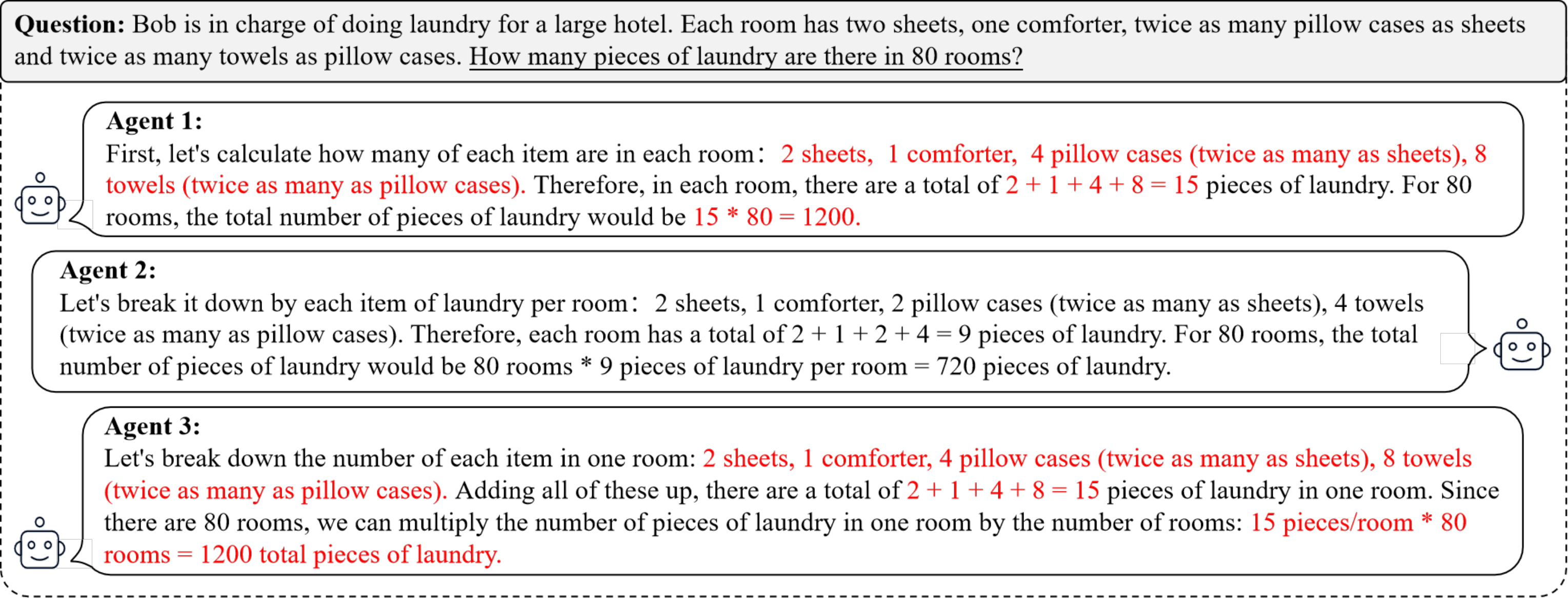}
     \caption{\textbf{Redundant Viewpoints Exchange between Agents.} The perspectives of Agent 1 and Agent 3 demonstrate a notable similarity. Throughout the debate, these viewpoints are exchanged with Agent 2, who receives these akin and repetitive viewpoints.
 }
     \label{fig:examples}
     \vspace{-1.0 em}
 \end{figure*}
 
To address these challenges, researchers have introduced techniques such as Chain-of-Thought (CoT) reasoning \citep{wei2023chainofthought} which decomposes complex problems into sequential steps, and self-consistency (SC) mechanisms \citep{wang2023selfconsistency}, along with self-correction strategies \citep{welleck2022generating,madaan2024self,shinn2024reflexion}. 

Despite these innovations, studies have shown that LLMs still struggle to improve through self-correction alone \citep{huang2023large, valmeekam2023can, stechly2023gpt}.
An emerging alternative is the Multi-agent Debate (MAD) framework, in which multiple independent agents propose and critique their own answers through rounds of debate, ultimately converging on a more robust consensus \citep{sun2023corex}. 
MAD has demonstrated promise in addressing the limitations of LLM self-correction by leveraging diverse agent perspectives to refine answers over iterative discussions \citep{chan2023chateval,du2023improving,liang2023encouraging}. 
However, as the number of agents and debate rounds increase, the token cost escalates significantly, limiting the scalability of MAD, especially in resource-constrained environments \citep{li2024improving,liu2024GroupDebate}.
To alleviate the token cost problem in multi-agent debates, researchers have proposed several strategies. 
For instance, \citep{du2023improving} summarizes agents' outputs at the end of each round, while \citep{sun2023corex} introduces a "forgetting" mechanism, where only the previous round’s outputs is retained for future rounds. 
Another approach, Sparse-MAD (S-MAD) \citep{li2024improving}, reduces communication overhead by limiting information exchange to neighboring agents. 
GroupDebate (GD) \citep{liu2024GroupDebate} further reduces token cost by clustering agents into smaller debate groups that exchange intermediate results between groups.

Although the reduction in token cost have achieved by the aforementioned approaches, our experiment reveals a substantial presence of redundancy and duplicate information in the inter-agent information exchange. As depicted in Figure \ref{fig:examples}, Agent 1 and Agent 3 exhibit repetitive viewpoints, leading to exacerbate the issue of token cost due to redundant duplication during the inter-agent information exchange. The issue of redundancy and duplication primarily stems from two potential factors: the limited solution space inherent in complex reasoning tasks, and the tendency of large language models to generate repetitive responses when faced with similar inputs \citep{holtzman2019curious,xu2022learning,yan2023understanding}.

\begin{figure*}[t]
    \centering
    \includegraphics[width=1\linewidth]{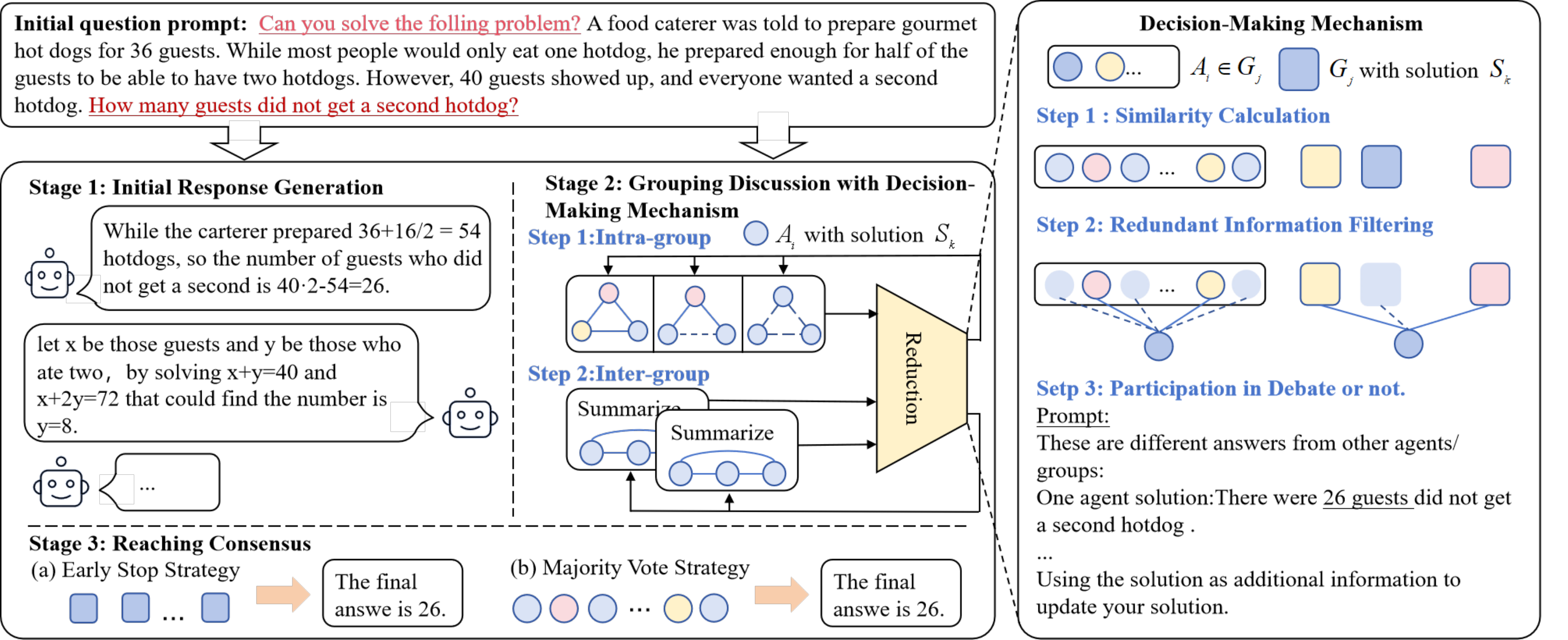}
     \caption{\textbf{Process of \({\text{S}^2\text{-MAD}}\).} The \({\text{S}^2\text{-MAD}}\) includes three stages: all agents generate initial responses independently at the first round and participate in group discussions to reach consensus under a Decision-Making Mechanism, which comprises: (1) Similarity calculation module accesses the similarity of responses either between or within groups. (2) Redundancy filter module filters redundant information, retaining only unique information that differs from the agent's own perspective. (3) Conditional participation module decide to participate in debate or not.
}
    \label{fig:framework}
    \vspace{-1.0 em}
\end{figure*}

To address these limitations, we propose a novel approach \textbf{S}elective \textbf{S}parse \textbf{M}ulti-\textbf{A}gent \textbf{D}ebate (\({\text{S}^2\text{-MAD}}\)), as shown in Figure \ref{fig:framework}. This approach utilizes a Decision-Making Mechanism to determine whether to participate in the debate, thereby further reducing token cost within multi-agent debates. Specifically, based on a grouping strategy, \({\text{S}^2\text{-MAD}}\) first generates initial viewpoints for the agents. In each round of debate, the Decision-Making Mechanism enables agents to selectively incorporate non-redundant responses that differ from their current viewpoints for answer checking and updating. The agents have the option to selectively engage in both intra-group and inter-group discussions, enabling them to actively participate in debates. The process concludes either when consensus is reached among the agents or when a final answer is obtained through majority voting.

To validate the effectiveness of \({\text{S}^2\text{-MAD}}\), we conduct a theoretical analysis of total token cost and perform extensive experiments across five tasks using different models. These experiments compare \({\text{S}^2\text{-MAD}}\) with existing multi-agent debate strategy as well as single-agent reasoning approaches, demonstrating its capability to significantly reduce token counts while maintaining comparable accuracy. Specifically, \({\text{S}^2\text{-MAD}}\) reduces token cost by up to 94.5\% compared to MAD, 90.2\% compared to MAD-Sparse, and 87.0\% compared to GD, while also significantly reducing token cost by up to 81.7\% compared to CoT-SC.  Importantly, these reductions come with a performance degradation of less than 2.0\%, demonstrating that \({\text{S}^2\text{-MAD}}\) maintains high accuracy while minimizing communication overhead. 

The main contributions of this paper are as follows:
\begin{enumerate}
    \item We propose \({\text{S}^2\text{-MAD}}\), an innovative sparse multi-agent debate strategy with Decision-Making Mechanism that reduces redundant information and inefficient debate. 
    \item We theoretically demonstrate the token cost advantages of \({\text{S}^2\text{-MAD}}\) over MAD, S-MAD, and GD.
    \item We validate the effectiveness of \({\text{S}^2\text{-MAD}}\) across five datasets using commercial and open-source models, demonstrating a significant reduction in token cost with minimal performance loss.


\end{enumerate}

\section{Preliminary} 
\label{sec:preliminary}

\paragraph{Problem Definition.} MAD, which integrates multiple agents for interactive communication to derive solutions, has demonstrated an effective approach in the application of LLMs, particularly in addressing complex logical reasoning and mathematical problems \citep{liang2023encouraging, chan2023chateval}. Given an input question $Q$ that requires an answer, a total of $M$ participating Language Model (LLM)-based agents engage in a multi-round debate, which is denoted as $A_i$, where $i \in \{1, 2, \ldots, M\}$.
Given a total of $T$ debate rounds, each round of debate is denoted as $t \in \{1, 2, \ldots, T\}$. We define the output of each agent $A_i$ at round $t$ as $O_i^{t}$. We assume that the upper bound of the token cost for each agent's output is $C$. Our goal is to maximize answer accuracy while minimizing token consumption through optimizing the interaction patterns among the agents in multi-agent debate.

\paragraph{MAD-based Methods and Token Cost.}
(i) MAD \citep{liang2023encouraging} involves several steps. Initially, each agent is provided with a question and generates an individual response. These responses then form the new input context for each agent, leading to the generation of subsequent responses. This debate procedure is repeated over multiple rounds, with the final answer derived through majority voting. The token cost complexity is $Token^{MAD} = \mathcal{O}\left( MTQ+(M^2T+MT^2)C \right)$. (ii) S-MAD \citep{li2024improving} decreases token consumption by sparsifying the fully connected topology of information exchange among agents within the standard MAD framework. Let $P_r$ denote the probability of each edge being removed. The token cost complexity can be represented as $Token^{S-MAD} = \mathcal{O}\left( MTQ+(1-P_r)M^2TC \right)$. (iii) GD \citep{liu2024GroupDebate} reduces token consumption through a group discussion strategy. Let $N$ denote the number of groups and $S$ represent the number of inter-group debate stages. The token cost complexity is computed as $Token^{GD} = \mathcal{O}\left(MTQ+(\frac{M^2T} {N}+MSN)C\right)$.
\section{Methodology}

\label{sec:methodology}

In this section, we first introduce the overall process of \({\text{S}^2\text{-MAD}}\) along with the details of Decision-Making Mechanism. And then we provide mathematical analysis of the token cost for our method subsequently.

\subsection{Selective Sparse MAD Process} \label{}

As illustrated in Figure \ref{fig:framework}, the debate process of \({\text{S}^2\text{-MAD}}\) consists of three main stages: the generation of initial responses, group discussions under the Decision-Making Mechanism and finally reaching consensus. 

\paragraph{Initial Response Generation.} In the initial round of debate, each agent is initialized as a LLM. To simulate diverse thought processes and ensure the generation of varied solutions, we employ a random decoding strategy by adjusting the temperature of model. During the first round, all agents independently produce their respective solutions for the given problem. 


\paragraph{Grouping Discussion with Decision-Making Mechanism.} From the second round onward, the Decision-Making Mechanism empowers agents to evaluate whether to engage in debate by assessing the similarity of intra- or inter-group viewpoints relative to their own perspectives. Agents will actively participate in debates when they encounter responses that present differing viewpoints, either from within their group or from other groups. Following these discussions, agents update their answers accordingly based on insights gained during the debate process. 

\paragraph{Reaching Consensus.} Our approach incorporates an early termination mechanism that allows us to conclude the debate when information has been exchanged between groups and all summarized viewpoints align. Conversely, if discrepancies remain among agents' solutions after the debate concludes, a majority vote will determine which solution is accepted as consensus.


\subsection{Decision-Making Mechanism} Upon receiving information, the Decision-Making Mechanism first employs the Similarity Calculation Module to calculate similarities among different pieces of information. Subsequently, it eliminates redundant perspectives of agents through the Redundancy Filtering Module. Finally, Conditional Participation Module is utilized to determine whether the agent should engage in the debate.


\paragraph{Similarity Calculation Module.} Following the generation of outputs, each agent undertakes a comprehensive assessment of the similarity between its own output and those produced by other agents or groups. This evaluation can be conducted through various methodologies; in this context, we employ a straightforward approach that involves analyzing key points within the outputs to determine their degree of similarity. Specifically, we employ regular expression matching to extract answers from the agents' responses and identical answers are considered to reflect similar viewpoints. Additionally, we also propose an alternative vectorization-based approach, where the responses are vectorized using an embedding model, and the cosine similarity is computed to evaluate the similarity of their viewpoints. In our further experiments, we conduct a comprehensive comparison of the performance of these two methods (See Section \ref{sec:in-depth-analysis}). By focusing on essential elements, agents can effectively gauge how closely aligned their perspectives are with those presented by others.

\paragraph{Redundancy Filtering Module.} Prior to engaging in the debate, agents systematically filter all incoming information to ensure relevance and uniqueness. Outputs that are identified as similar to either their own or previously received viewpoints are promptly discarded from consideration. This rigorous filtering process guarantees that each agent exclusively considers unique perspectives during discussions, thereby minimizing redundancy and fostering a more dynamic exchange of ideas.

\paragraph{Conditional Participation Module.} Agents actively engage in debate when divergent viewpoints exist within or among groups, recognizing that such differences enrich the discourse and lead to more robust conclusions. Conversely, if all outputs align consistently without variation, agents will opt to remain silent rather than contribute redundant information. At the conclusion of each round of debate, agents update their knowledge base with accepted viewpoints gleaned from interactions with others; this iterative learning process enhances their ability to respond thoughtfully and effectively in subsequent rounds.

\subsection{Token Cost Analysis} \label{token-consumption-analysis}




In \({\text{S}^2\text{-MAD}}\), we summarize the outputs from whtnin each group at the end of each stage. 
Given a group of agents $G_j$ which has completed a stage $s$ of debate, we denote its summary as $Sum_j^s$.
Since in \({\text{S}^2\text{-MAD}}\)  each agent determines participation based on whether viewpoints are consistent, we define the number of agents with differing viewpoints from the $i^{th}$ agent $D_i$ is:

\begin{equation}
\begin{split}
\left\{
\begin{aligned}
&\sum_{i'\in G_j} Sim(O_i^{t-1} ,O_{i'}^{t-1})<\epsilon,\\
&\qquad (s-1)R+1<t<min(sR,T)\\
&\sum_{1}^{N} Sim(O_i^{t-1},Sum_{j}^{s-1})<\epsilon,\\
&\qquad\qquad\qquad\qquad t=(s-1)R+1
\end{aligned}
\right.
\end{split}
\end{equation}
Therefore, apart from generating the initial answer, the probability of agent $A_i$ participating in the debate in round $t$ is
\begin{equation}
\begin{split}
P_i^t=
\left\{
\begin{aligned}
&1, && D^t_i>0\\  
&0, && D^t_i=0
\end{aligned}
\right.
\end{split}
\end{equation}
Then token cost $Token^t_s$ in round $t$ at stage $s$ is:
\begin{equation}
\sum_{j=1}^N\sum_{i\in G_j} P_i^t(Q+O_i^t+\frac{MD_i^t}{N}\sum_{i'\in G_j}O_{i'}^{t-1})
\end{equation}
where $(s-1)R + 1 < t <= min(sR, T)$,  and 
\begin{equation}
    \sum_{i=1}^MP_i^t(Q+O_i^{t-1}+O_i^t+\frac{D_i^t}{N}\sum_{j=1}^NSum_j^{s-1})
\end{equation}
where $t = (s-1)R + 1$. Finally, the total token cost of \({\text{S}^2\text{-MAD}}\) is $Token = \mathcal{O}\left(MTQ+(\frac{M^2T}{N}+MSN)CP\right)$, where $C$ represents the upper bound on the token number for each agent's response and the generated summary, $P$ represents the upper bound of the average probability of each agent participating in the debate globally. More calculation details are shown in Appendix \ref{appendix:token-cost-S^2-MAD}.

\paragraph{Discussion.} From the perspective of total token cost complexity comparison, \({\text{S}^2\text{-MAD}}\) exhibits the same token cost complexity as standard MAD since the initial viewpoints of agents are generated by retaining the question input. However, for the same $M$ and $T$, since agents’ answers tend to become consistent as the debate progresses, we define the probability of obtaining a answer different with other agents is $p$. Thus, the token cost will only increase to be comparable to that of Group Debate when different answers are obtained in each round, which occurs with a probability of only $p^{MN}$.

\section{Experiments}
\label{sec:experiments}

\begin{table*}[t]
\centering
\setlength\tabcolsep{1.5 pt}
\resizebox{\textwidth}{!}{
    \begin{tabular}{ccccccccccc} 
    \toprule
    \multirow{2}{*}{Methods} & \multicolumn{2}{c}{GSM8K} & \multicolumn{2}{c}{MATH} & \multicolumn{2}{c}{MMLU} & \multicolumn{2}{c}{GPQA} & \multicolumn{2}{c}{Arithmetic}  \\ 
    \cmidrule{2-11}
    & ACC(\%)$\uparrow$  & Tokens($k$)$\downarrow$   & ACC(\%)$\uparrow$   & Tokens($k$)$\downarrow$   & ACC(\%)$\uparrow$   & Tokens($k$)$\downarrow$       & ACC(\%)$\uparrow$    & Tokens($k$)$\downarrow$      & ACC(\%)$\uparrow$   & Tokens($k$)$\downarrow$    \\ 
    \midrule
    \rowcolor{Gray}
    \multicolumn{11}{c}{\texttt{GPT-3.5-turbo-0125}} 
    \\
    \midrule
    CoT                      & 78.8\scriptsize{$\pm 0.04$}  & 0.25\scriptsize{$\pm 0.03$}             & 35.2\scriptsize{$\pm 0.02$} & 0.37\scriptsize{$\pm 0.01$}         & 72.9\scriptsize{$\pm 0.02$} & 0.24\scriptsize{$\pm 0.00$}                & 31.2\scriptsize{$\pm 0.03$}  & 2.02\scriptsize{$\pm 0.02$}         & 82.2\scriptsize{$\pm 0.04$} & 0.16\scriptsize{$\pm 0.02$}           \\
    CoT-SC(40)               & \underline{\textbf{85.6}}\scriptsize{$\pm 0.01$} & 10.0\scriptsize{$\pm 0.02$}          & \underline{\textbf{48.2}}\scriptsize{$\pm 0.01$} & 14.6\scriptsize{$\pm 0.15$}        & \underline{\textbf{78.4}}\scriptsize{$\pm 0.01$}  & 9.64\scriptsize{$\pm 0.03$}                 & 32.0\scriptsize{$\pm 0.01$}  & 80.5\scriptsize{$\pm 0.27$}         & 95.0\scriptsize{$\pm 0.01$} & 6.25\scriptsize{$\pm 0.13$}           \\
    MAD(5,4)                 & 83.6\scriptsize{$\pm 0.01$} & 20.4\scriptsize{$\pm 0.12$}          & 40.5\scriptsize{$\pm 0.00$}  & 23.4\scriptsize{$\pm 0.24$}        & 74.1\scriptsize{$\pm 0.03$} & 26.9\scriptsize{$\pm 0.11$}                & 41.4\scriptsize{$\pm 0.03$} & 64.3\scriptsize{$\pm 3.06$}         & 96.2\scriptsize{$\pm 0.02$}  & 20.3\scriptsize{$\pm 0.32$}           \\ 
    S-MAD(5,4)                 & \underline{85.6}\scriptsize{$\pm 0.02$} & 17.9\scriptsize{$\pm 0.22$}          & 40.3\scriptsize{$\pm 0.01$}  & 19.0\scriptsize{$\pm 0.17$}        & 74.3\scriptsize{$\pm 0.01$} & 22.2\scriptsize{$\pm 0.23$}                & \underline{45.0}\scriptsize{$\pm 0.00$} & 57.3\scriptsize{$\pm 0.77$}         & 96.8\scriptsize{$\pm 0.01$}  & 9.83\scriptsize{$\pm 0.06$}           \\ 
    GD(5,4)                  & 85.4\scriptsize{$\pm 0.03$} & 15.5\scriptsize{$\pm 0.08$}          & \underline{40.7}\scriptsize{$\pm 0.01$} & 18.7\scriptsize{$\pm 0.13$}         & 76.0\scriptsize{$\pm 0.02$} & 16.1\scriptsize{$\pm 0.06$}                & \underline{\textbf{45.0}}\scriptsize{$\pm 0.03$} & 50.5\scriptsize{$\pm 0.67$}         & \underline{99.3}\scriptsize{$\pm 0.00$} & 13.7\scriptsize{$\pm 0.64$}           \\ 
    \({\text{S}^2\text{-MAD(5,4)}}\)                  & 84.8\scriptsize{$\pm 0.02$} & \underline{4.53}\scriptsize{$\pm 0.31$}          & 40.3\scriptsize{$\pm 0.00$} & \underline{11.4}\scriptsize{$\pm 0.29$}         & \underline{76.8}\scriptsize{$\pm 0.02$} & \underline{4.78}\scriptsize{$\pm 0.57$}                & 43.8\scriptsize{$\pm 0.01$} & \underline{23.6}\scriptsize{$\pm 7.05$}         & \underline{\textbf{99.6}}\scriptsize{$\pm 0.00$} & \underline{2.29}\scriptsize{$\pm 0.07$}           \\
    \midrule
    \rowcolor{Gray}
    \multicolumn{11}{c}{\texttt{GPT-4-0613}} 
    \\
    \midrule
    CoT                      & 92.8\scriptsize{$\pm 0.01$} & 0.38\scriptsize{$\pm 0.00$}           & 73.0\scriptsize{$\pm 0.01$} & 0.71\scriptsize{$\pm 0.00$}         & 83.7\scriptsize{$\pm 0.01$} & 0.39\scriptsize{$\pm 0.00$}                & 47.2\scriptsize{$\pm 0.02$}  & 2.23\scriptsize{$\pm 0.10$}         & - & -          \\
    CoT-SC(40)               & \underline{\textbf{94.3}}\scriptsize{$\pm 0.00$} & 15.2\scriptsize{$\pm 0.03$}          & \underline{\textbf{81.0}}\scriptsize{$\pm 0.01$}  & 27.9\scriptsize{$\pm 0.00$}         & 88.4\scriptsize{$\pm 0.01$}     & 15.4\scriptsize{$\pm 0.02$}                & 53.0\scriptsize{$\pm 0.00$} & 90.0\scriptsize{$\pm 0.30$}         & - & -           \\
    MAD(5,4)                 & 93.3\scriptsize{$\pm 0.01$} & 50.4\scriptsize{$\pm 0.19$}          & 78.7\scriptsize{$\pm 0.00$} & 70.9\scriptsize{$\pm 0.20$}        & \underline{90.8}\scriptsize{$\pm 0.00$} & 61.7\scriptsize{$\pm 0.09$}                & 59.7\scriptsize{$\pm 0.01$} & 109.6\scriptsize{$\pm 5.22$}         & - & -           \\ 
    S-MAD(5,4)                 & 93.0\scriptsize{$\pm 0.02$} & 24.4\scriptsize{$\pm 0.04$}          & \underline{79.3}\scriptsize{$\pm 0.00$}  & 59.5\scriptsize{$\pm 0.53$}        & \underline{\textbf{91.5}}\scriptsize{$\pm 0.00$} & 48.2\scriptsize{$\pm 0.24$}                 & \underline{60.7}\scriptsize{$\pm 0.02$} & 97.8\scriptsize{$\pm 0.42$}         & -  & -           \\ 
    GD(5,4)                  & \underline{94.3}\scriptsize{$\pm 0.00$} & 21.4\scriptsize{$\pm 0.08$}           & 76.7\scriptsize{$\pm 0.01$} & 36.8\scriptsize{$\pm 0.28$}         & 87.8\scriptsize{$\pm 0.01$} & 25.3\scriptsize{$\pm 0.11$}                & \underline{\textbf{62.7}}\scriptsize{$\pm 0.02$} & 64.6\scriptsize{$\pm 1.25$}
    & - & - \\
    \({\text{S}^2\text{-MAD(5,4)}}\)                  & 94.2\scriptsize{$\pm 0.00$} & \underline{2.78}\scriptsize{$\pm 0.16$}          & 77.3\scriptsize{$\pm 0.01$} & \underline{11.2}\scriptsize{$\pm 0.79$}         & 88.1\scriptsize{$\pm 0.01$} & \underline{4.71}\scriptsize{$\pm 0.35$}                & 60.8\scriptsize{$\pm 0.04$} & \underline{27.1}\scriptsize{$\pm 9.5$}         & - & -           \\
    \midrule
    \rowcolor{lightgray}
    \rowcolor{Gray}
    \multicolumn{11}{c}{\texttt{Llama-3.1-8B-Instruct}} 
    \\
    \midrule
    CoT                      & 83.2\scriptsize{$\pm 0.01$} & 0.32\scriptsize{$\pm 0.00$}           & 32.0\scriptsize{$\pm 0.03$} & 0.53\scriptsize{$\pm 0.00$}         & 61.2\scriptsize{$\pm 0.04$} & 0.43\scriptsize{$\pm 0.00$}         & 19.7\scriptsize{$\pm 0.02$}                & 2.35\scriptsize{$\pm 0.01$}         & 74.0\scriptsize{$\pm 0.01$} & 0.19\scriptsize{$\pm 0.00$}          \\
    CoT-SC(40)               & \underline{\textbf{89.0}}\scriptsize{$\pm 0.01$} & 12.6\scriptsize{$\pm 0.00$}          & 43.0\scriptsize{$\pm 0.01$}  & 21.1\scriptsize{$\pm 0.02$}         & \underline{74.1}\scriptsize{$\pm 0.02$}     & 17.2\scriptsize{$\pm 0.04$}                & 33.5\scriptsize{$\pm 0.02$} & 93.9\scriptsize{$\pm 0.01$}         & 83.0\scriptsize{$\pm 0.01$} & 7.63\scriptsize{$\pm 0.00$}           \\
    MAD(5,4)                 & 86.7\scriptsize{$\pm 0.02$} & 31.4\scriptsize{$\pm 0.27$}          & \underline{\textbf{46.0}}\scriptsize{$\pm 0.01$} & 80.5\scriptsize{$\pm 0.20$}        & 73.4\scriptsize{$\pm 0.02$} & 54.7\scriptsize{$\pm 0.58$}                & 37.0\scriptsize{$\pm 0.01$} & 117.5\scriptsize{$\pm 4.23$}         & \underline{\textbf{91.0}}\scriptsize{$\pm 0.02$} & 76.1\scriptsize{$\pm 0.12$}           \\ 
    S-MAD(5,4)                 & \underline{87.3}\scriptsize{$\pm 0.02$} & 26.0\scriptsize{$\pm 0.36$}          & \underline{45.0}\scriptsize{$\pm 0.00$}  & 66.3\scriptsize{$\pm 0.43$}        & \underline{\textbf{74.5}}\scriptsize{$\pm 0.00$} & 43.3\scriptsize{$\pm 0.14$}                 & \underline{\textbf{40.0}}\scriptsize{$\pm 0.01$} & 102.2\scriptsize{$\pm 2.53$}         &  89.5\scriptsize{$\pm 0.01$} & 62.3\scriptsize{$\pm 0.53$}           \\ 
    GD(5,4)                  & 86.5\scriptsize{$\pm 0.00$} & 17.0\scriptsize{$\pm 0.07$}           & 44.0\scriptsize{$\pm 0.01$} & 39.9\scriptsize{$\pm 0.28$}         & 73.5\scriptsize{$\pm 0.01$} & 39.5\scriptsize{$\pm 0.11$}                & 37.0\scriptsize{$\pm 0.02$} & 71.6\scriptsize{$\pm 2.25$}
    & 89.3\scriptsize{$\pm 0.03$} & 33.4\scriptsize{$\pm 0.02$}  \\
    \({\text{S}^2\text{-MAD(5,4)}}\)                  & 85.7\scriptsize{$\pm 0.02$} & \underline{5.39}\scriptsize{$\pm 0.16$}          & 44.0\scriptsize{$\pm 0.05$} & \underline{21.9}\scriptsize{$\pm 0.17$}         & 73.8\scriptsize{$\pm 0.04$} & \underline{10.6}\scriptsize{$\pm 1.74$}                & \underline{39.0}\scriptsize{$\pm 0.04$} & \underline{19.3}\scriptsize{$\pm 1.5$}         & \underline{90.0}\scriptsize{$\pm 0.03$} & \underline{13.9}\scriptsize{$\pm 0.29$}          \\
    \bottomrule
    \end{tabular}
    }
    \caption{\textbf{Comparison of Token Cost and Accuracy Between \({\text{S}^2\text{-MAD}}\) and Other Methods.} The results of highest accuracy are \textbf{bold} and the results of both highest accuracy and lowest token cost except from CoT are \underline{underlined}. The dash (-) indicates that the model achieved a correctness rate of 1 for all methods on this dataset.}
\label{tab:comparison_all}
\end{table*}

\subsection{Experimental Setup} 
\paragraph{Tasks and Metrics.} To evaluate the effectiveness and efficiency of \({\text{S}^2\text{-MAD}}\) in mathematical and logical reasoning tasks, we use total token cost and accuracy (ACC) as evaluation metrics across five representative tasks: (1) GSM8K \cite{cobbe2021training}: a dataset designed to assess the model's reasoning ability in complex mathematical problems. (2) MATH \cite{hendrycks2021measuring}: a dataset covers various branches of mathematics to evaluate the capacity to generate problem-solving logic and reasoning processes. (3) MMLU \cite{hendrycks2020measuring}: a dataset that aimed at evaluating the model's overall performance across diverse tasks. (4) GPQA \cite{rein2023gpqa}: a multiple-choice question dataset,  containing 448 questions across various disciplines. (5) Arithmetic \cite{brown2020language}: a datasets evaluates the model's fundamental mathematical reasoning abilities.

\paragraph{Baselines.} We compare our \({\text{S}^2\text{-MAD}}\) with the following baselines: (1) Chain-of-Thought (CoT) \cite{wei2023chainofthought}; (2) Self-Consistency with Chain-of-Thought (CoT-SC) \cite{wang2023selfconsistency}; (3) Multi-agent Debate (MAD) \cite{liang2023encouraging}; (4) Sparse MAD (S-MAD) \cite{li2024improving}; (5) GroupDebate (GD) \cite{liu2024GroupDebate}.Experiments are conducted with different numbers of agents, rounds, and group strategies. For example, (5,4) represents using 5 agents and 4 rounds, while CoT-SC(40) indicates CoT-SC with 40 reasoning paths.

\paragraph{Implementation Details.} We set the number of intra-group rounds to 2 and use a forgetting mechanism to retain the outputs from the previous round only. At the end of each intra-group discussion phase, we filter and summarize the results from the same groups. Our experiments use GPT-3.5-turbo-0301, GPT-4-0613 and Llama-3.1-8B-Instruct as agents, evaluating all baselines and our \({\text{S}^2\text{-MAD}}\) in a zero-shot setting. Since the accuracy rate of the Arithmetic dataset reached 100\% in a single GPT-4, no further comparison was conducted. Details about the prompts 
and additional results for GPT-4o-mini and GPT-4o-0806 are showed in the Appendix \ref{appendix:prompts} and Appendix \ref{sec:experiments_add}.

For the Similarity Calculation Module, we primarily use regular expression matching for the main results and cosine similarity for further analysis, which uses the Bert-base-uncased model to vectorize the agent’s responses and calculate the cosine similarity between the responses. 

\subsection{Main Result} 
In this section, we conducted a detailed comparison of our method with multi-agent debate methods (including MAD, S-MAD, GD) and single-agent methods (including CoT, CoT-SC). The main observations are as follows:
firstly, we compare \({\text{S}^2\text{-MAD}}\) with MAD. The results presented in Table \ref{tab:comparison_all} shows that \({\text{S}^2\text{-MAD}}\) consistently reduces total token cost across different models while maintaining comparable accuracy, it achieves a reduction of 94.5\%, 84.2\%, 92.4\%, 83.6\% and 88.7\% on the five datasets respectively compared to MAD. The variation in these percentages is due to the varying difficulty of the questions, which impacts model performance. Furthermore, compared to S-MAD and GD, our approach achieves up to 90.2\% and 87.0\% less token cost, respectively. This demonstrates that there is a significant amount of redundancy in the information exchange during multi-agent debate, leading to the inefficiency of token cost throughout the debate process.

We also conducted a comparison with the single-agent method CoT, achieving a significant improvement in accuracy across five datasets, especially achieving up to 19.3\% and 12.6\% on GPQA and MMLU dataset. Furthermore, when compared with the Cot-SC method, we successfully reached or exceeded Cot-SC's performance on certain datasets, such as GPQA and Arithmetic, while using relatively fewer token cost. 

\begin{table}[h]
  \centering
  \setlength\tabcolsep{1.5 pt}
  \begin{tabular}{lccc}
    \toprule
    \textbf{Method} & \textbf{ACC (\%)} & \textbf{Token (k)} & \textbf{Cost Saving}\\ 
    \midrule
    MAD(5,4) & \underline{72.3}\scriptsize{$\pm 0.00$} & 78.7\scriptsize{$\pm 0.31$} & - \\
    \midrule
    \({\text{S}^2\text{-MAD}}\) & &\\
    \midrule
    RE-Matching & 70.7\scriptsize{$\pm 0.01$} & \underline{12.4}\scriptsize{$\pm 0.73$} & -84.2\%\\
    VecCS$_{\tau = 0.96}$ & 69.0\scriptsize{$\pm 0.02$} & 18.6\scriptsize{$\pm 0.67$} & -76.4\%\\
    VecCS$_{\tau = 0.40}$ & \textbf{74.5}\scriptsize{$\pm 0.02$} & \textbf{4.18}\scriptsize{$\pm 0.09$} & -94.7\%\\
    \bottomrule
    
  \end{tabular}
  \caption{\textbf{Comparison of different similarity calculation strategies on MATH using GPT-4o-mini.} RE-Matching refers to regular expression matching and VecCS$_{\tau = 0.96}$ means vectorization and cosine similarity calculation with $\tau = 0.96$. The results of highest accuracy or lowest token cost are \textbf{bold} and the suboptimal results are \underline{underlined}.}
  \label{tab:comparison_similarity}
\end{table}
\subsection{In-Depth Analysis} 
\label{sec:in-depth-analysis}

\begin{figure}[t]
  \includegraphics[width=\columnwidth]{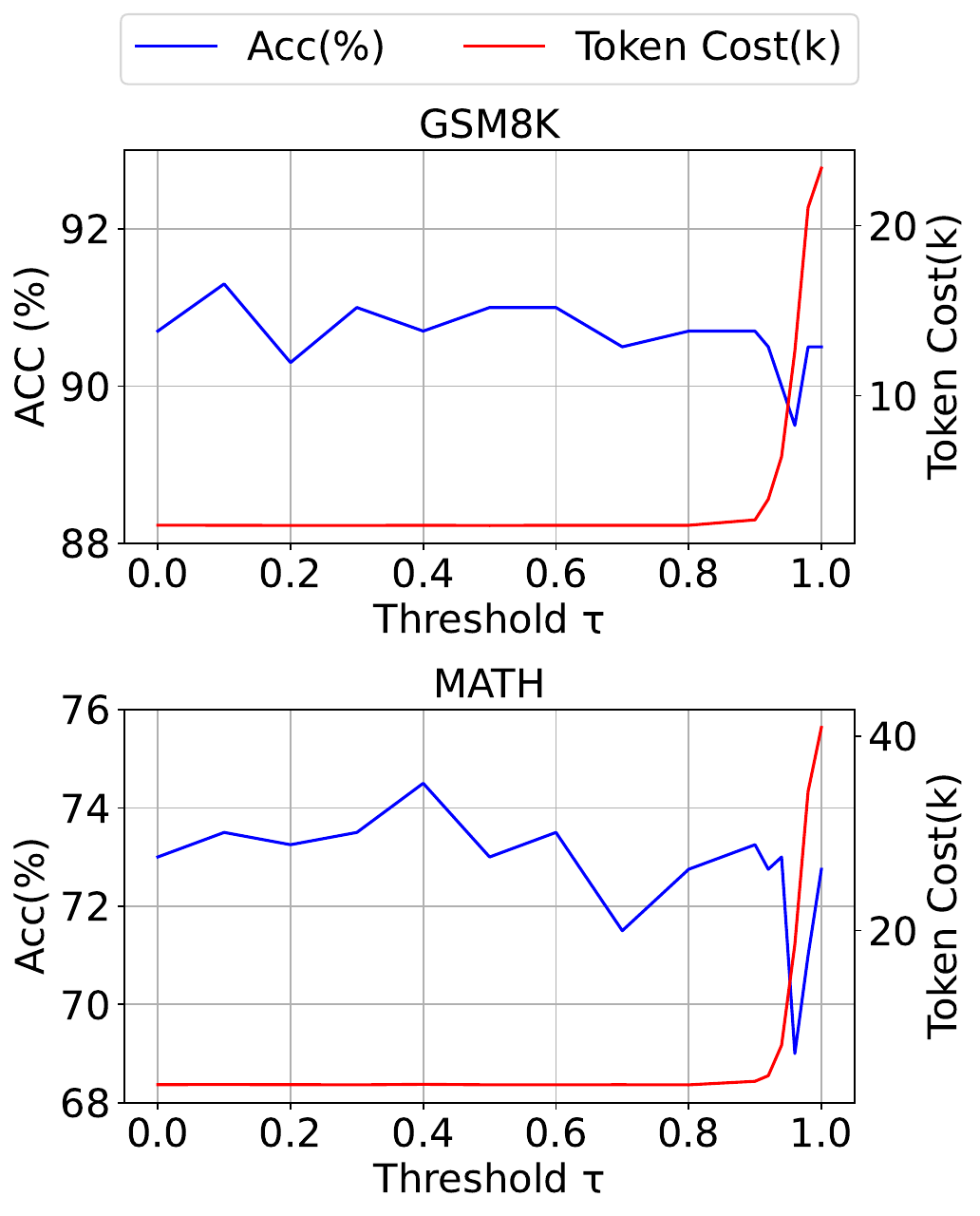}
    \caption{\textbf{The relationship between the threshold} $\tau$, ACC, and Token Cost on the GSM8K and MATH datasets.}
    \label{fig:vect}
    \vspace{-1.0 em}
\end{figure}

\paragraph{Similarity Calculation Strategy.} In this section, we conduct further comparison on similarity calculation strategies using GPT-4o-mini. As shown in Table \ref{tab:comparison_similarity}, the method of vectorizing the responses and calculating their cosine similarity can achieve the best accuracy and the lowest token cost at a specific threshold. Specifically, on the MATH dataset using GPT-4o-mini, setting $\tau$ to 0.40 results in a 2.2\% improvement in accuracy and a 94.7\% reduction in token cost compared to the MAD method. However, when $\tau$ is set to 0.96, the increased token cost actually leads to a decrease in ACC. Furthermore, as illustrated in the Figure \ref{fig:vect}, the token cost remains relatively low when $\tau < 0.85$, but increases sharply thereafter. This is attributed to the prompt's strict formatting constraints on the agent's output, which cause high similarity among outputs. Additionally, we observed that the relative optimal threshold values for accuracy vary across different datasets (e.g., approximately 0.1 for GSM8K and 0.4 for MATH), making it challenging to manually determine the optimal threshold settings.

\begin{table}[h]
  \centering
  \setlength\tabcolsep{1.5 pt}
  \begin{tabular}{lccc}
    \toprule
    \textbf{Method} & \textbf{ACC (\%)} & \textbf{Token (k)} & \textbf{Cost Saving}\\ 
    \midrule
    MAD(8,3) & 86.7\scriptsize{$\pm 0.02$} & 28.5\scriptsize{$\pm 0.08$} & - \\
    S-MAD(8,3) & 86.5\scriptsize{$\pm 0.01$} & 18.7\scriptsize{$\pm 0.04$}  & -34.4\%\\
    \midrule
    GD & &\\
    \midrule
    2+6 & 86.7\scriptsize{$\pm 0.00$} & 20.3\scriptsize{$\pm 0.08$} & -28.8\%\\
    4+4 & 87.3\scriptsize{$\pm 0.01$} & 19.1\scriptsize{$\pm 0.00$}  & -33.0\%\\
    2+3+3 & 87.3\scriptsize{$\pm 0.02$} & 18.0\scriptsize{$\pm 0.05$} & -36.8\%\\
    2+2+4 & \underline{87.7}\scriptsize{$\pm 0.00$} & 18.3\scriptsize{$\pm 0.03$} & -35.7\%\\
    2+2+2+2 & \underline{\textbf{87.8}}\scriptsize{$\pm 0.00$} & 17.4\scriptsize{$\pm 0.04$} & -38.9\%\\
    \midrule
    \({\text{S}^2\text{-MAD}}\) & &\\
    \midrule
    2+6 & 84\scriptsize{$\pm 0.00$} & 8.01\scriptsize{$\pm 0.13$} & -71.9\%\\
    4+4 & 84.6\scriptsize{$\pm 0.00$} & 7.28\scriptsize{$\pm 0.19$} & -74.5\%\\
    2+3+3 & 85.1\scriptsize{$\pm 0.00$} & \underline{6.97}\scriptsize{$\pm 0.14$} & -75.5\%\\
    2+2+4 & 84.5\scriptsize{$\pm 0.00$} & 7.02\scriptsize{$\pm 0.20$} & -75.4\%\\
    2+2+2+2 & 83.4\scriptsize{$\pm 0.02$} & \underline{\textbf{6.78}}\scriptsize{$\pm 0.12$} & -76.2\%\\
    \bottomrule
    
  \end{tabular}
  \caption{\textbf{Comparison of different group strategies with GD and \({\text{S}^2\text{-MAD}}\) on GSM8K datasets.} The notation 2+6 signifies two distinct groups containing 2 and 6 agents respectively. The results of highest accuracy or lowest token cost are \textbf{bold} and the suboptimal results are \underline{underlined}.}
  \label{tab:comparison}
\end{table}

\paragraph{Group Strategy.} To assess the impact of different grouping strategies on performance and token cost, we conducted experiments involving 8 agents across 3 rounds on the GSM8K. As shown in Table \ref{tab:comparison}, increasing the number of groups can reduce the total token cost as the quantity of information exchange is limited by communication constraints. However, when the number of agents within a group increases, agents can more effectively receive diverse information, achieving higher accuracy.  Our findings indicate a clear trade-off between optimizing token cost and maintaining high accuracy, emphasizing the importance of selecting an appropriate grouping strategy.

\begin{figure}[t]
  \includegraphics[width=\columnwidth]{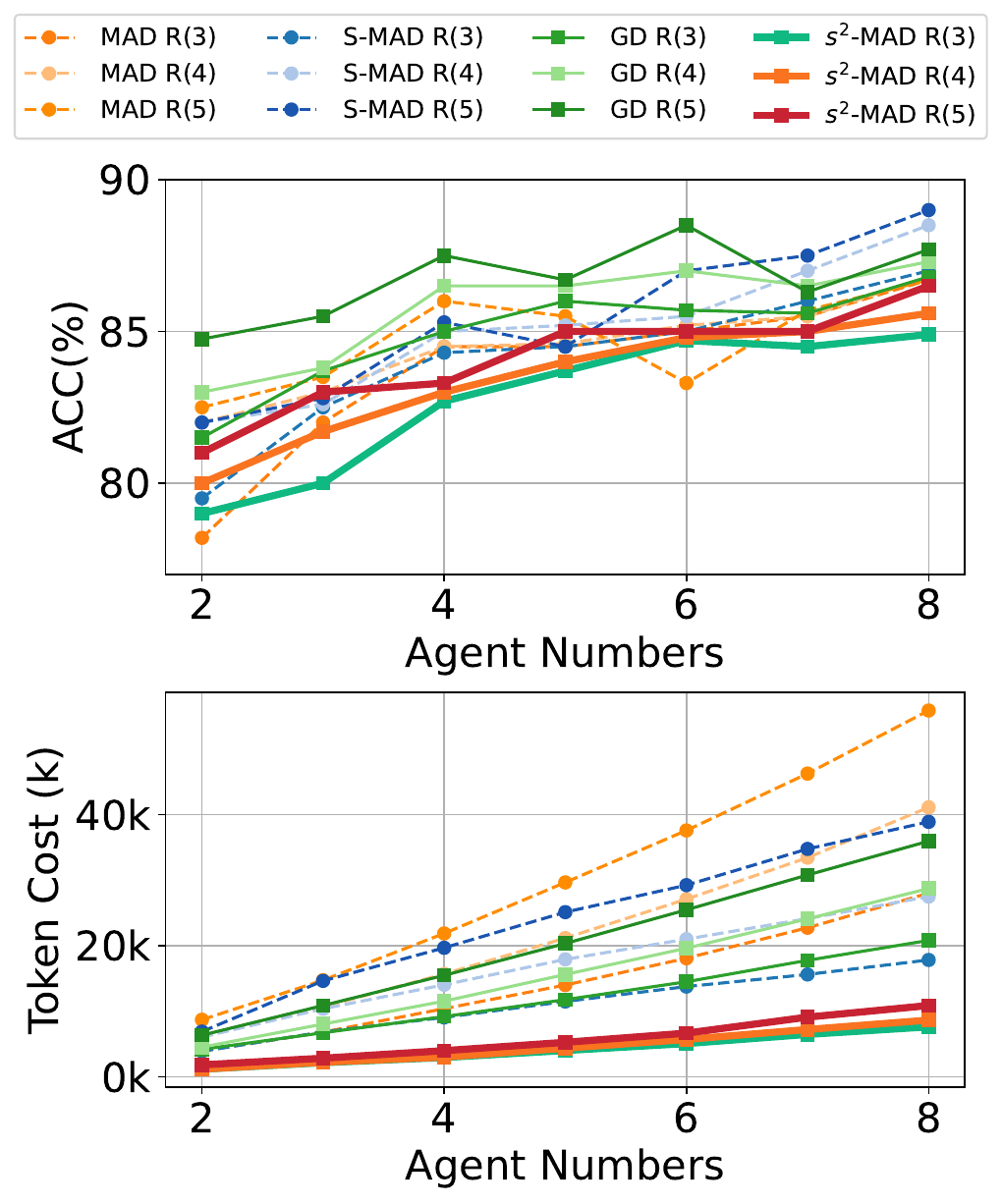}
    \caption{\textbf{Scaling study of Agents and Rounds.}}
    \label{fig:e_2}
    \vspace{-1.0 em}
\end{figure}

\begin{figure*}[t]
    \centering
    \includegraphics[width=1\linewidth]
{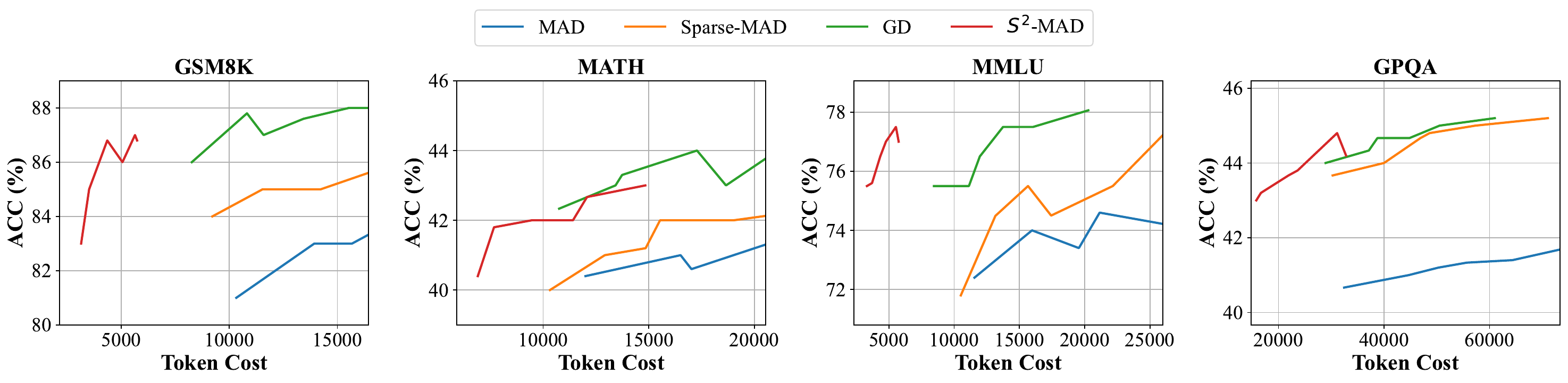}
    \caption{\textbf{Scaling Study of Token Cost.}}
    \label{fig:scale}
    \vspace{-1.0 em}
\end{figure*}

\paragraph{Agent, Round and Token Cost Scaling.} To assess the impact of the number of rounds and agents on the accuracy and token cost across different methods, we analyze the trends in accuracy and token cost for different combinations of rounds and agents. As shown in the Figure \ref{fig:e_2}, with the increase in the number of agents and rounds, there is a noticeable enhancement in the overall performance of various methods; however, this also leads to a significant increase in token cost. Our approach maintains a certain level of performance while exhibiting a gradual increase in token cost as agents and rounds increase, achieving the lowest token cost across different setting, as shown in Figure \ref{fig:scale}. This indicates that there is a significant amount of redundant and repetitive information exchange during debates, resulting in higher token cost and less effective agent interactions.

\subsection{Ablation Study} 
To investigate the impact of different modules and strategies on performance, we conducted ablation experiments on the GSM8K dataset using GPT-3.5-turbo, are shown in Table \ref{tab:ablation}. Our \({\text{S}^2\text{-MAD}}\) achieved an accuracy of 85.6\% with a token cost of 4.73k, demonstrating a significant 72.7\% reduction compared to MAD. We further explored sparsity through constrained communication topologies, which slightly decreased accuracy to 84.7\% while retaining a similar token cost. Without the early stopping strategy led to a slight accuracy drop to 84.4\% but maintained a comparable token cost of 4.99k. In contrast, removing the jump strategy resulted in a more substantial decline in accuracy to 80.8\% and an increase in token usage to 9.45k, We hypothesize that this is due to insufficient information diversity, causing redundant checks that impact response accuracy. Finally, although removing the filtering module can increase accuracy to 87.6\%, it also leads to an increase in token cost of 13.4k. Although our method has not yet achieved optimal performance in terms of accuracy and token cost, it still shows a slight improvement over the MAD method while significantly reducing token usage, highlighting the efficiency of our proposed method in balancing accuracy and token saving.

\section{Related Work}
\label{sec:related-work}

\subsection{LLM Reasoning}
Many studies have explored ways to improve the logical reasoning abilities of LLMs. CoT~\citep{wei2023chainofthought} mimics human thought processes by breaking complex tasks into sequential steps.
Many CoT variants~\citep{eric2022star,wang2023selfconsistency, shum2023automatic} extend this framework by generating multiple reasoning chains and selecting the optimal one based on specific criteria.
Building on this, Tree-of-Thoughts (ToT)~\citep{yao2024tree} structures the reasoning process into a tree-like path, where each step serves as a decision point, enabling the evaluation of multiple reasoning paths and self-assessment. 
Similarly, Skeleton-of-Thought~\citep{ning2023skeleton} accelerates answer generation by first creating a skeletal framework and then completing the content in parallel for each point. 
Table-of-Thoughts~\citep{jin2023tab} improves reasoning accuracy through structured modeling of the reasoning process.
While these CoT-based methods follow structured reasoning paths, more complex reasoning structures have been proposed. For instance, Graph-of-Thoughts~\citep{besta2024graph} models reasoning as a flexible graph, allowing for non-linear task solving beyond the limitations of chains or trees. Methods such as Least-to-Most~\citep{zhou2022least} and Lambada~\citep{kazemi2022lambada} take a problem decomposition approach, breaking tasks into subproblems and solving them step-by-step, where each sub-answer informs the next step.
Additionally, frameworks like LReasoner~\citep{wang2021logic} introduce mechanisms that enhance reasoning by extracting logical structures embedded in the problem. Logic-LM~\citep{pan2023logic} combines symbolic solvers to convert natural language into symbolic formulas and introduces a self-refinement module to correct errors during the reasoning process. 
%
\begin{table}[t]
    \centering
    \setlength\tabcolsep{1.5 pt}
    \begin{tabular}{lcc}
        \toprule
        \textbf{Method} & \textbf{ACC (\%)} & \textbf{Token (k)} \\ 
        \midrule
        MAD & 85.4\scriptsize{$\pm 0.02$} & 18 \\ 
        \midrule
        \({\text{S}^2\text{-MAD}}\) & 85.6\scriptsize{$\pm 0.00$} & 4.73 (-72.7\%) \\ 
        \midrule
        w/ Sparse Commu. & 84.7\scriptsize{$\pm 0.00$} & 4.71 (-73.8\%) \\  
        w/o Early Stop& 84.4\scriptsize{$\pm 0.00$} & 4.99 (-72.3\%) \\ 
        w/o Jump & 80.8\scriptsize{$\pm 0.02$} & 9.45 (-47.5\%) \\ 
        w/o Filter & 87.6\scriptsize{$\pm 0.01$} & 13.4 (-25.6\%) \\ 
        \bottomrule
    \end{tabular}
    \caption{\textbf{Comparison of accuracy and cost saving against MAD on GSM8K dataset.} All experiments were conducted using GPT-3.5-turbo.}
    \label{tab:ablation}
\end{table}
\subsection{Multi-agent Debate}
MAD is a promising approach to enhance the reasoning capabilities of LLMs by facilitating discussions among multiple agents who collaboratively refine and update generated answers. 
\cite{liang2023encouraging} presents a MAD framework where multiple agents engage in "tit for tat" argumentation, managed by a judge, to stimulate divergent thinking in LLMs.
Building on this foundation, ~\cite{xiong2023examining} introduce the FORD framework, which organizes a three-stage debate aligned with real-world scenarios, comprising fair debate, mismatched debate, and round-table debate formats.
%
~\cite{xu2023toward} present a framework that mirrors the academic peer review process, allowing models to autonomously develop solutions, review each other’s work, and revise their answers based on feedback. 
ChatEval~\citep{chan2023chateval}, another MAD framework, employs diverse communication strategies and varied role prompts to foster human-like interactions and evaluations in natural language dialogue. 
Moreover, ~\cite{wang2023apollo} address cognitive constraints in multi-agent debates by integrating prior knowledge retrieval and a self-selection module, enhancing reasoning capabilities and overall performance.
Further exploring collaboration, ~\cite{fu2023improving} analyze the autonomous enhancement of negotiation strategies among LLMs through role-playing and iterative AI feedback within a structured negotiation game, highlighting the trade-offs between deal quality and risk management.
However, as the number of agents and debate rounds increases, token costs can rise significantly. 
To mitigate this, \cite{du2023improving} suggests summarizing agent outputs at the end of each round for subsequent inputs, and \cite{sun2023corex} introduces a "forgetfulness" mechanism to retain only the previous round's output. The MAD-Sparse approach \cite{li2024improving} utilizes a sparse communication strategy, limiting information exchange to adjacent agents. Additionally, GroupDebate \cite{liu2024GroupDebate} promotes a grouping strategy, allowing agents to debate internally while sharing interim results. However, these methods do not enable agents to critically assess the redundancy of incoming information, limiting overall efficiency.

\section{Conclusion}
\label{sec:conclusion}
In this work, we identified the issue of redundant viewpoints among agents in Multi-agent Debate (MAD). To address this, we proposed \textbf{S}elective \textbf{S}parse \textbf{M}ulti-\textbf{A}gent \textbf{D}ebate (\({\text{S}^2\text{-MAD}}\)), a novel strategy designed to reduce token cost by selectively incorporating non-redundant viewpoints from different agents, thereby significantly improving the efficiency of information exchange and debate.
Our theoretical analysis verify the effectiveness of $\text{S}^2\text{-MAD}$, and extensive experiments conducted on five benchmark datasets demonstrate that $\text{S}^2\text{-MAD}$ can significantly reduce token cost in MAD while maintaining competitive performance.
For future work, we aim to refine $\text{S}^2\text{-MAD}$ by further optimizing the identification and condensation of non-redundant viewpoints between agents, with the goal of further reducing token cost and enhancing efficiency. Additionally, exploring methods to increase the diversity of thought among agents will be key to improving the overall accuracy of $\text{S}^2\text{-MAD}$.
\section{Limitation}
\label{sec:limitation}
Despite the significant reduction in token cost achieved by $\text{S}^2\text{-MAD}$, our method has several limitations. First, the reduction in token cost exhibits variability depending on the consistency of agent responses. When agents' answers differ significantly, the efficiency gains are limited, whereas more consistent responses yield a greater reduction in token cost. This variability introduces an element of unpredictability to the system's overall efficiency.
Second, the judge module in $\text{S}^2\text{-MAD}$ is sometimes unable to filter out redundant viewpoints. The module relies on keyword extraction using regular expressions to determine whether agents' outputs convey the same idea. However, when agents express similar views with different wording or synonyms, the judge module may fail to detect these similarities, resulting in redundant exchanges of information. This can undermine the potential gains in efficiency and contribute to token cost redundancy.
Therefore, there remains room for improvement in optimizing the token cost of $\text{S}^2\text{-MAD}$. 

\section*{Acknowledgments}
This work was supported in part by the National Natural Science Foundation of
China(NSFC) with Grant No. 62172383 and No. 62231015, Anhui Provincial Key
R\&D Program with Grant No. S202103a05020098, Research Launch Project of University of Science and Technology of China(USTC) with Grant No. KY0110000049.

\bibliography{custom}
\appendix
\section{Selective Sparse MAD Algorithm} \label{appendix:s^2-al}
The detailed \({\text{S}^2\text{-MAD}}\) Algorithm is as follows:
\begin{breakablealgorithm}
    \caption{\({\text{S}^2\text{-MAD}}\) Methods}
    \label{alg:S^2_MAD}
    \begin{algorithmic}[1]
        \Require Number of groups $N$, number of agents $M$, question $Q$, total rounds $T$, intra-group debate round $R$, total stages $S$, redundancy filter $F$, Opinion Judger $J$, answer extracter $VOTE$ 
        \Ensure $Answer$
     
        \State $A \gets [A_1,A_2,\ldots,A_M]$ 
        \Statex\Comment{Initialize and shuffle the agents randomly}
        \State $G \gets [G_1,G_2,\ldots,G_N]$  
        \Statex\Comment{Initialize each group} 
        \State $H \gets [H_1,H_2,\ldots,H_M]$  
        \Statex\Comment{Initialize each agent with empty memory}
        
        \State $Sum \gets [Sum_1, Sum_2,\ldots,Sum_N]$  

        \For{$i=1$ to $M$}
            \State $H_i \gets [Q]$ 
        \EndFor
        
        \For{$s=1$ to $S$}
            \For{$j=1$ to $N$}
                \For{$t = (s-1)R+1$ to $\min(sR,T)$}
                    \For{$A_i \in G_j$}
                        \If{$s=1$ \textbf{and} $t=1$}
                            \State $h_i \gets A_i(H_i)$ 
                            \State $H_i \gets H_i + h_i$ 
                            \State $H_i \gets H_i + BUF$ 
                        \Else
                            \State $buf \gets [\quad]$
                            \If{$s \neq 1$ \textbf{and} $t=(s-1)R+1$}
                                \For{$S_i \in Sum$}
                                    \If{$J(H_i[-2],S_i)$}
                                        \State $buf \gets buf + S_i$ 
                                    \EndIf
                                \EndFor
                            \Else
                                \For{$A_{i'} \in G_j$ \textbf{and} $A_{i'} \neq A_i$}
                                    \If{$J(H_i[-2],H_{i'}[-2])$}
                                        \State $buf \gets buf + Replay_{i'}$  
                                    \EndIf
                                \EndFor
                            \EndIf
                            \If{$len(buf) \ne 0$}
                                \State $H_i[-1] \gets buf$ 
                                \State $h_i \gets A_i(H_i)$ 
                                \State $H_i[-2] \gets h_i$   
                            \EndIf
                        \EndIf
                    \EndFor
                \EndFor
                \If{$s \neq S$}
                    \State $summary \gets [\;]$
                    \For{$A_i \in G_j$}                       
                        \State $sum \gets sum + H_i[-2]$         
                    \EndFor
                    \State $Sum_j \gets LLM(sum)$ 
                \EndIf 
            \EndFor
            \State $Sum \gets F(Sum)$
            \If{$len(Sum)=1$}
                break\Comment{End debate if only one summary}
            \EndIf                
        \EndFor
        \State $Answer \gets VOTE(H)$
        
    \State \Return $Answer$
    \end{algorithmic}
\end{breakablealgorithm}

\section{Token Cost Analysis} \label{appendix:token-cost-S^2-MAD}
In this appendix section, we aim to provide a theoretical analysis of the token cost for \({\text{S}^2\text{-MAD}}\). As LLMs' outputs typically are not too long and we can actually control the token length of LLMs' outputs in prompts to some extent, we assume that the upper bound on the number of tokens output by each agent participating in debate is $Output_{max}$ and the upper bound on the number of tokens in the generated summary is $Summary_{max}$. We define $C$ as the maximum of $Output_{max}$ and $Summary_{max}$, $P$ represents the upper bound of the average probability of each Agent participating in the debate globally.

As mentioned in Section \ref{sec:methodology}, our \({\text{S}^2\text{-MAD}}\) includes three types of processes and thus the total token cost $Token^{GD}$ can be further dividied into:
\begin{equation}
\begin{aligned} 
    &Token = \underbrace{Token_1^1}_\text{initial thinking} + \\ 
    &\underbrace{\sum_{s=2}^{S}(Token^{summary}_{s-1}+Token_s^{(s-1)R+1})}_\text{inter-group discussion} \\
    &+ \underbrace{\sum_{s=1}^{S}\sum_{t=(s-1)R+2}^{min(sR,T)}Token_s^t}_\text{intra-group discussion}
\end{aligned}
\end{equation}\label{eq:divide}
Specifically, for initial thinking, the token cost of each agent includes the initial question prompt and its own output. For intra-group debate, the token cost of each agent includes the unique responses from other agents within the same group that differ from its own in the previous round and its output. For inter-group debate, the token cost includes the summary generation cost, which comprises the unique responses from all groups and the output summary, as well the token cost of each agent which comprises as its output and summary from from other groups that differ from its own. The detailed computation process of the token cost in \({\text{S}^2\text{-MAD}}\) can be found in Algorithm \ref{alg:token_cost_s^2_mad}.

\begin{algorithm}[ht]
    \caption{Tokens Cost in \({\text{S}^2\text{-MAD}}\) Methods}
    \label{alg:token_cost_s^2_mad}
    \begin{algorithmic}[1]
        \Require Number of groups $N$, number of agents  $M$, question length $Q$, total rounds $T$, group debate round $R$, total stages $S$, summary of each group at the end of each stage $Summary = \{Summary_j^s|j=1,2, \ldots,N,s=1,2,\ldots,S\}$ , output length of each agent $A_i(i=1,2, \ldots,M)$ in each round $t(t=1,2,\ldots,T)$ $Output_i^t$,  each group agents set $G = \{G_j|j=1,2, \ldots,N\}$, probability of participating in the debate of each agent $P=\{P_i^t|i=1,2,\ldots,M,t=1,2,\ldots,T\}$.
        \Ensure Total token cost $Token$
        
        \State $Token_1^{1} \gets M \times Q + \sum_{i=1}^{M} O_i^1$ \\
        \Statex\Comment{First round} 
        
        \For{$t = 2$ to $R$}
            \State $Token_1^t \gets \sum_{j=1}^{N} \sum_{i \in G_j} (Q  + O_i^{t}+$
            \Statex $\qquad \qquad \qquad \qquad \qquad \sum_{i' \in G_j} O_{i'}^{t-1})$ 
            \Statex \Comment{Subsequent rounds of the first stage}
        \EndFor
        
        \For{$s=2$ to $S$}
            \State $Token_{s-1}^{summary} \gets \sum_{j=1}^{N} (\sum_{i \in G_j}$
            \Statex $\qquad \qquad O_i^{(s-1)R} + Sum_j^{s-1})$ 
            \Statex \Comment{Summary at the end of stage $s-1$}
            
            \State $Token_s^{(s-1)R+1} \gets \sum_{i=1}^{M} P_i^{(s-1)R+1}(Q+$
            \Statex $\qquad \quad O_i^{(s-1)R}+ \sum_{j=1}^{N} Sum_j^{s-1}$
            \Statex $\qquad \quad + O_i^{(s-1)R+1})$ 
            \Statex \Comment{First round of the stage $s$}
        
            \For{$t = (s-1)R+2$ to $\min(sR,T)$}
                \State $Token_s^t \gets \sum_{j=1}^{N} \sum_{i \in G_j} P_i^t(Q  + $
                \Statex $\qquad \qquad O_i^{t} + \sum_{i' \in G_j} O_{i'}^{t-1})$ 
                \Statex \Comment{Subsequent rounds of the stage $s$}
            \EndFor
        \EndFor
        
        \State $Token \gets  \sum_{t=1}^{R}Token_1^t+\sum_{s=2}^S($
        \Statex $Token_{s-1}^{summary}+\sum_{t=(s-1)R+1}^{\min(sR,T)}Token_s^t)$ 
        \Statex \Comment{Total token cost in debate}
        
        \State \Return $Token$
    \end{algorithmic}
\end{algorithm}

Following Alogorithm~\ref{alg:token_cost_s^2_mad} and Eq.~\ref{eq:divide}, we have: 
\begin{equation}
\begin{aligned}
    &Token = MQ + \sum_{i=1}^M O_i^1 \\
    &+ \sum_{s=2}^S [\sum_{j=1}^N(\sum_{i\in G_j} O_i^{(s-1)R}+Sum_j^{s-1}) \\
    &+\sum_{i=1}^MP_i^{(s-1)R}(Q+O_i^{(s-1)R+1}\\
    &\qquad+\frac{D_i^t}{N}\sum_{j=1}^N Sum_j^{s-1} +Output_i^{(s-1)R+1})]\\
    &+\sum_{s=1}^S \sum_{t=(s-1)R+2}^{min(sR,T)}\sum_{j=1}^N\sum_{i\in G_j}P_i^t(Q+O_i^t\\
    &\qquad\qquad\qquad\qquad\qquad+\frac{MD_i^t}{N}\sum_{i' \in G_j}O_{i'}^{t-1}) \\
    & \leq MTQ + P_{max} \times \{\\
    &\quad[3MS-2M+(T-S)(K+1)M] \times O_{max}\\
    &\quad+ (S-1)(M+1)N\times Sum_{max} \}\\
    & \leq MTQ+P_{max} \times \{\frac{2M^2T}{N}\times O_{max} \\
    &\quad+ 2MSN\times Sum_{max}\} \\
    & = \mathcal{O}\left(MTQ+(\frac{M^2T}{N}+MSN)CP\right)
\end{aligned}
\end{equation}

When we set $N \rightarrow \mathcal{O}\left(\sqrt{\frac{MT}{S}}\right)$, we can theoretically obtain $Token \rightarrow \mathcal{O}\left(MTQ+\sqrt{M^3TS}CP\right)$. Furthermore, If we consider setting $S$ to a very small positive integer and the average probability of their participation decreases as the capability of individual agents improves, then $Token$ can approach $\mathcal{O}\left(MTQ+\sqrt{M^3T}CP\right)$. This complexity is significantly lower than that of MAD.

\section{Prompts} \label{appendix:prompts}
In this section, we present some examples of prompts. Table \ref{tab:input-prompts} displays the input prompts used in our \({\text{S}^2\text{-MAD}}\) across different datasets, which encompass five different types. Table \ref{tab:Requirements} outlines the prompts regarding output Format Requirements in our \({\text{S}^2\text{-MAD}}\).

\begin{table*}[t]
\setlength{\tabcolsep}{5.5pt}
\centering
\resizebox{\textwidth}{!}{
\begin{tabular}{ccl}
    \toprule
    {\bf Type} & {\bf Task} & {\bf Prompt} \\
    \midrule
     \multirow{4}{*}{System}  & \multirow{4}{*}{All} & \emph{Welcome to the debate! You are a seasoned debater with expertise in succinctly and persuasively expressing your viewpoints. }\\
     & & \emph{You will be assigned to debate groups, where you will engage in discussions with fellow participants. The outcomes of }\\
     & & \emph{each group's deliberations will be shared among all members. It is crucial For you to leverage this inFormation effectively }\\
     & & \emph{in order to critically analyze the question at hand and ultimately arrive at the correct answer. Best of luck!} \\
     \midrule
      \multirow{5}{*}{Starting}  & \multirow{1}{*}{Arithmetic} & \emph{What is the result of \{\}+\{\}*\{\}+\{\}-\{\}*\{\}? $<$ Output Format $>$.}   \\
      \cmidrule{2-3}
      & \multirow{1}{*}{GSM8K} & \emph{Can you solve the following math problem? $<$Problem$>$ Explain your reasoning.  $<$ Output Format $>$.}\\
      \cmidrule{2-3}
     & \multirow{1}{*}{MMLU} & \emph{Can you answer the following question? $<$Problem$>$: A) {}, B) {}, C) {}, D) {} Explain your answer, $<$Output Format$>$.} \\
     \cmidrule{2-3}
     & \multirow{1}{*}{MATH} & \emph{Can you solve the following math problem? $<$Problem$>$ Explain your reasoning as concise as possible. $<$Output Format$>$ . } \\
     \cmidrule{2-3}
     & \multirow{1}{*}{GPQA} & \emph{Can you answer the following question? $<$Problem$>$: A) {}, B) {}, C) {}, D) {} Explain your answer, $<$Output Format$>$.} \\
     \midrule
      \multirow{2}{*}{Intra-group Debate}  & \multirow{2}{*}{All} & \emph{These are the recent unique opinions from other agents that differ with yours: $<$other agent responses$>$ Using the opinions} \\
      & & \emph{carefully as additional advice, can you provide an updated answer?}\\
      && \emph{Examine your solution and that other agents step by step. $<$Output Format$>$ .} \\
     \midrule
     \multirow{3}{*}{Summary}  & \multirow{3}{*}{All} & \emph{These are the recent/updated and unique opinions from all agents: $<$all agent responses$>$}\\
     & & \emph{Summarize these opinions carefully and completly in no more than 80 words. }\\
     & & \emph{Aggregate and put your final answers in parentheses at the end of your response. }\\
     \midrule
     \multirow{3}{*}{Inter-group Debate}  & \multirow{3}{*}{All} & \emph{These are the recent unique opinions from all groups: one group responses: $<$group summary$>$. } \\
      & & \emph{Using the reasoning from all groups as additional advice, can you give an updated answer?} \\
      & & \emph{Examine your solution and that all groups step by step. $<$Output Format$>$.} \\
    \bottomrule
\end{tabular}
}
\caption{\textbf{Prompts in Each Stage.} List of prompts used in each task. }
\label{tab:input-prompts}
\end{table*}

\begin{table*}[t]
\centering
\resizebox{\textwidth}{!}{
\begin{tabular}{@{\extracolsep{2pt}}lc@{}}
\toprule
Dataset & \multicolumn{1}{c}{Output Format Requirements} \\
\midrule
\multirow{1}{*}{Arithmetic} & \emph{Make sure to State your answer at the end of the response.}\\
\midrule
\multirow{2}{*}{GSM8K} & \emph{Your final answer should be a single numerical number, in the Form \textbackslash boxed\{\{answer\}\},}\\
& \emph{at the end of your response.}\\
\midrule
\multirow{1}{*}{MMLU} & \emph{Put your final choice in parentheses at the end of your response.}\\
\midrule
\multirow{1}{*}{MATH} & \emph{Put your final answer in the Form \textbackslash boxed\{\{answer\}\}, at the end of your response.}\\
\midrule
\multirow{1}{*}{GPQA} & \emph{Put your final answer in the Form \textbackslash The correct answer is (insert answer here)}\\
\bottomrule
\end{tabular}
}
\caption{\textbf{Output Format Requirements in Each Dataset.}}
\label{tab:Requirements}
\end{table*}

\newpage

\section{More Result}

\label{sec:experiments_add}

\begin{table*}[t]
\centering
\setlength\tabcolsep{1.5 pt}
\resizebox{\textwidth}{!}{
    \begin{tabular}{ccccccccc} 
    \toprule
    \multirow{2}{*}{Methods} & \multicolumn{2}{c}{GSM8K} & \multicolumn{2}{c}{MATH} & \multicolumn{2}{c}{MMLU} & \multicolumn{2}{c}{GPQA}   \\ 
    \cmidrule{2-9}
    & ACC(\%)$\uparrow$  & Tokens($k$)$\downarrow$   & ACC(\%)$\uparrow$   & Tokens($k$)$\downarrow$   & ACC(\%)$\uparrow$   & Tokens($k$)$\downarrow$       & ACC(\%)$\uparrow$    & Tokens($k$)$\downarrow$   \\ 
    \midrule
    \rowcolor{Gray}
    \multicolumn{9}{c}{\texttt{GPT-4o-mini}} 
    \\
    \midrule
    MAD(5,4)                 & \underline{91.0}\scriptsize{$\pm 0.00$} & 50.6\scriptsize{$\pm 0.16$}          & \underline{72.3}\scriptsize{$\pm 0.00$}  & 78.7\scriptsize{$\pm 0.31$}        & \underline{89.5}\scriptsize{$\pm 0.02$} & 63.4\scriptsize{$\pm 0.29$}                & 43.7\scriptsize{$\pm 0.00$} & 118.9\scriptsize{$\pm 2.33$}                    \\ 
    S-MAD(5,4)                 & 90.0\scriptsize{$\pm 0.01$} & 50.6\scriptsize{$\pm 0.49$}          & 71.7\scriptsize{$\pm 0.02$}  & 65.9\scriptsize{$\pm 0.42$}        & 89.1\scriptsize{$\pm 0.01$} & 49.3\scriptsize{$\pm 0.19$}                & 44.7\scriptsize{$\pm 0.02$} & 97.6\scriptsize{$\pm 0.30$}             \\ 
    GD(5,4)                  & 89.3\scriptsize{$\pm 0.00$} & 22.1\scriptsize{$\pm 0.07$}          & 72.0\scriptsize{$\pm 0.01$} & 38.8\scriptsize{$\pm 0.07$}         & 88.8\scriptsize{$\pm 0.00$} & 23.9\scriptsize{$\pm 0.08$}                & \underline{46.6}\scriptsize{$\pm 0.00$} & 64.9\scriptsize{$\pm 0.48$}                    \\ 
    \({\text{S}^2\text{-MAD(5,4)}}\)                  & 90.7\scriptsize{$\pm 0.01$} & \textbf{3.29}\scriptsize{$\pm 0.20$}          & 70.7\scriptsize{$\pm 0.01$} & \textbf{12.4}\scriptsize{$\pm 0.73$}         & 86.1\scriptsize{$\pm 0.00$} & \textbf{3.84}\scriptsize{$\pm 0.28$}                & 42.3\scriptsize{$\pm 0.02$} & \underline{24.5}\scriptsize{$\pm 7.38$}            \\
    \({M\text{S}^2\text{-MAD(5,4)}}\) & \textbf{91.7}\scriptsize{$\pm 0.01$} & \underline{3.75}\scriptsize{$\pm 0.19$}          & \textbf{72.3}\scriptsize{$\pm 0.02$} & \underline{14.4}\scriptsize{$\pm 0.10$}         & \textbf{89.8}\scriptsize{$\pm 0.01$} & \underline{5.50}\scriptsize{$\pm 0.39$}                & \textbf{46.8}\scriptsize{$\pm 0.03$} & \textbf{20.8}\scriptsize{$\pm 3.94$} \\
    \midrule
    \rowcolor{Gray}
    \multicolumn{9}{c}{\texttt{GPT-4o-0806}} 
    \\
    \midrule
    MAD(5,4)                 & \underline{94.0}\scriptsize{$\pm 0.00$} & 48.4\scriptsize{$\pm 0.08$}          & \textbf{79.0}\scriptsize{$\pm 0.01$} & 67.8\scriptsize{$\pm 0.09$}        & 88.4\scriptsize{$\pm 0.01$} & 52.9\scriptsize{$\pm 0.13$}                & 52.2\scriptsize{$\pm 0.04$} & 102.6\scriptsize{$\pm 2.84$}                    \\ 
    S-MAD(5,4)                 & 93.7\scriptsize{$\pm 0.00$} & 39.1\scriptsize{$\pm 0.15$}          & 76.7\scriptsize{$\pm 0.02$}  & 54.8\scriptsize{$\pm 0.11$}        & \underline{\textbf{89.8}}\scriptsize{$\pm 0.00$} & 41.2\scriptsize{$\pm 0.33$}                 & \textbf{53.0}\scriptsize{$\pm 0.01$} & 93.7\scriptsize{$\pm 2.16$}                    \\ 
    GD(5,4)                  & 92.7\scriptsize{$\pm 0.01$} & 20.7\scriptsize{$\pm 0.06$}           & 74.7\scriptsize{$\pm 0.01$} & 44.9\scriptsize{$\pm 0.07$}         & 88.4\scriptsize{$\pm 0.00$} & 22.4\scriptsize{$\pm 0.04$}                & 52.5\scriptsize{$\pm 0.03$} & 60.1\scriptsize{$\pm 1.34$}
     \\
    \({\text{S}^2\text{-MAD(5,4)}}\)                  & 92.8\scriptsize{$\pm 0.01$} & \textbf{2.93}\scriptsize{$\pm 0.20$}          & 75.3\scriptsize{$\pm 0.02$} & \textbf{11.8}\scriptsize{$\pm 0.46$}         & 88.3\scriptsize{$\pm 0.01$} & \textbf{4.34}\scriptsize{$\pm 0.13$}                & 51.0\scriptsize{$\pm 0.04$} & \textbf{21.9}\scriptsize{$\pm 7.49$}                 \\
    \({M\text{S}^2\text{-MAD(5,4)}}\) & \textbf{94.0}\scriptsize{$\pm 0.01$} & \underline{3.35}\scriptsize{$\pm 0.14$}          & \underline{77.0}\scriptsize{$\pm 0.01$} & \underline{12.7}\scriptsize{$\pm 1.41$}         & \underline{88.6}\scriptsize{$\pm 0.00$} & \underline{5.48}\scriptsize{$\pm 0.26$}                & \underline{52.7}\scriptsize{$\pm 0.01$} & \underline{26.5}\scriptsize{$\pm 9.82$} \\
    \bottomrule
    \end{tabular}
    }
    \caption{\textbf{Comparison of Token Cost and Accuracy Between \({\text{S}^2\text{-MAD}}\) and Other Methods.} The results of highest accuracy are \textbf{bold} and the results of both highest accuracy and lowest token cost except from CoT are \underline{underlined}. The dash (-) indicates that the model achieved a correctness rate of 1 for all methods on this dataset.}
\label{tab:comparison_add}
\end{table*}

In this appendix section, we conducted a detailed comparative experiment between our proposed method and other multi-agent debate methods using GPT-4o-mini and GPT-4o-0806. As shown in the Table \ref{tab:comparison_add}, \({\text{S}^2\text{-MAD}}\) consistently reduces the total token cost while maintaining comparable accuracy. Specifically, on four datasets, our method achieved a reduction of 94.3\%/84.2\%/94.0\%/79.4\% compared to MAD, respectively. Furthermore, to enhance the accuracy, we initialized agents with multiple prompt settings as \({\text{MS}^2\text{-MAD}}\), encouraging them to explore multiple thought paths, thereby achieving optimal accuracy. This suggests that promoting the exploration of multiple thought paths in multi-agent debates can be beneficial for the agent system to solve problems more accurately.

\end{document}